\documentclass[]{svproc} 

\usepackage{graphicx,amsmath,amssymb,bm,xcolor}
\usepackage[numbers]{natbib}
\usepackage{subfig}

\usepackage{url}

\newcommand{\trsp}{{\scriptscriptstyle\top}}
\newcommand{\psin}{{\dagger}}
\newcommand{\psinr}{{\ddagger}}

\newcommand{\ty}[1]{{\scriptscriptstyle{#1}}}

\definecolor{rr}{rgb}{.8,0,0}
\definecolor{gg}{rgb}{0,.7,0}
\definecolor{bb}{rgb}{0,0,.8}
\definecolor{rb}{rgb}{.4,0,.4}

\begin{document}
\mainmatter 

\title{Nullspace Structure in Model Predictive Control}

\author{Hakan Girgin \and Sylvain Calinon}
\institute{Idiap Research Institute, Martigny, Switzerland\\
\email{hakan.girgin@idiap.ch}, \email{sylvain.calinon@idiap.ch}\\
}

\maketitle

\begin{abstract}
Robotic tasks can be accomplished by exploiting different forms of redundancies. This work focuses on planning redundancy within Model Predictive Control (MPC) in which several paths can be considered within the MPC time horizon. We present the nullspace structure in MPC with a quadratic approximation of the cost and a linearization of the dynamics. We exploit the low rank structure of the precision matrices used in MPC (encapsulating spatiotemporal information) to perform hierarchical task planning, and show how nullspace computation can be treated as a fusion problem (computed with a product of Gaussian experts). We illustrate the approach using proof-of-concept examples with point mass objects and simulated robotics applications. 
\end{abstract}

\section{Introduction}
\label{sec:Intro}

\begin{figure}
  \centering
  \includegraphics[width=.7\columnwidth]{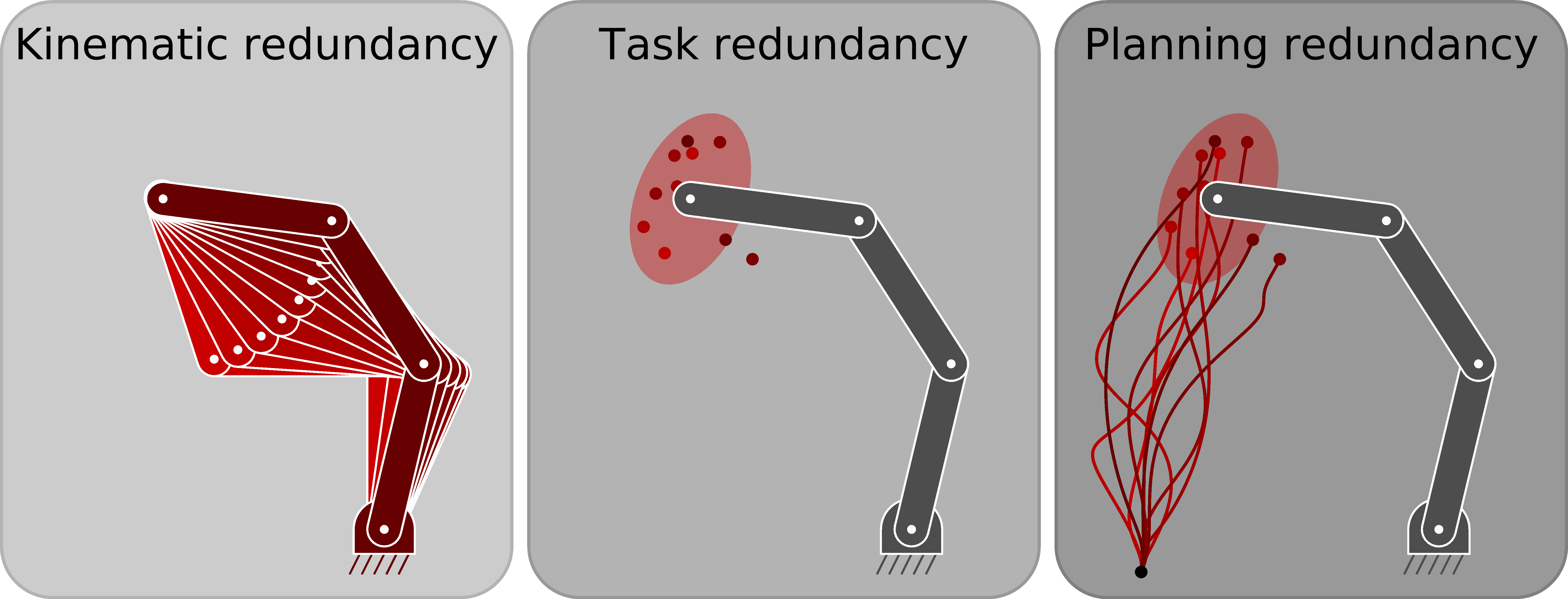}
  \caption{
\emph{From left to right:} kinematic/mechanical redundancy (morphology level), task redundancy (spatial level), planning redundancy (spatiotemporal level). 
  }
  \label{fig:redundancy}
\end{figure}

A collection of work in neuroscience informs us that being skillful is not related to being precise \cite{Todorov02,Wolpert11,Sternad10,Ganesh12}. It is instead related to the exploitation of various forms of redundancy and variations in an optimal way. Figure \ref{fig:redundancy} illustrates the various forms of redundancy that that can be exploited in robotics. In this planar example, kinematic redundancy arises when a 2D point is tracked by a 3-axis robot \cite{Walker88,Baker-88-IJRR,Siciliano90,Chiaverini08}. Most tasks do not require a specific point to be tracked, which means that rather than a point, tracking should instead consider a distribution of the different options in which the tip of the robot can move while satisfying the task constraints \cite{Calinon10IROS}. we show in this technical report that such redundancy can also be defined at a path level within an MPC formulation, by considering as an example the different options that a robot has to move through a sequence of viapoints.  

The structure of the technical report is as follows. In Section \ref{sec:IK} and \ref{sec:MPC}, we review inverse kinematics problem with nullspace structure and standard model predictive control with quadratic cost and linear dynamics, respectively. In Section \ref{sec:MPC_null}, we show how to apply the same principles of IK nullspace to MPC. We then show in Section \ref{sec:Experiments} some proof-of-concept examples with unit mass point agents.

\section{Inverse kinematics with nullspace structure}
\label{sec:IK}

In an inverse kinematic problem, the standard nullspace control formulation, with a velocity $\bm{\dot{x}}$ in Cartesian space as first task and a velocity $\bm{\dot{q}}$ in joint space as secondary task, can be found by solving in the control space $\bm{u}$ the constrained objective
\begin{align}
	\bm{\hat{u}} &= \arg\min_{\bm{u}} {\big\|\bm{u} - \bm{\dot{q}} \big\|}^2 
	\quad\mathrm{s.t.}\quad \bm{J}\bm{u}=\bm{\dot{x}}\nonumber\\
	&= \bm{J}^\psin \bm{\dot{x}} + \bm{N} \bm{\dot{q}},
	\label{eq:stdnull}
\end{align}
with a Jacobian matrix $\bm{J}$, its pseudoinverse $\bm{J}^\psin$ and the nullspace projection matrix $\bm{N} = \bm{I} - \bm{J}^\psin \bm{J}$ \cite{Liegeois77}, see Appendix \ref{app:nullspace} for details. Such nullspace computation can also be treated as a fusion problem (computed with a product of Gaussian experts), see Appendix \ref{app:nullspacePoG} for details.

The pseudoinverse $\bm{J}^\psin$ in \eqref{eq:stdnull} can be computed in different ways according to the rank of $\bm{J}$. If $\bm{J}$ is full row-rank, then $\bm{J}^\psin=\bm{J}^\trsp{(\bm{J}\bm{J}^\trsp)}^{-1}$, or else if $\bm{J}$ is full column-rank, then $\bm{J}^\psin = (\bm{J}^\trsp\bm{J})^{-1}\bm{J}^\trsp$. Using a singular value decomposition (SVD) to compute the pseudoinverse provides a more general method as it can be used either in rank-deficient case or in full-rank case. For simplicity, we will adopt in this paper the non-SVD perspective for the computation of pseudoinverses. Pseudoinverse computation details are given in Appendix \ref{app:nullspace}.

Standard inverse kinematics formulation with nullspace structure can be extended to have $K$ tasks with assigned priorities, given as
\begin{align}
	\bm{u} = \sum_{k = 1}^{K}   \prod_{i = 1}^{k}\bm{N}_{i-1}\bm{J}_k^\psin\bm{\dot{x}}_k,
	\label{eq:stdnullK}
\end{align}
where $\bm{N}_{i}$ is the nullspace matrix of $\bm{J}_{i}$ with $\bm{N}_0 = \bm{I}$.

A task-priority formulation \cite{Hanafusa32} describes a solution to the nullspace control of inverse kinematics problem as
\begin{align}
	\bm{u} = \bm{J}_{1}^\psin\bm{\dot{x}}_{1} + (\bm{J}_{2}\bm{N}_{1})^\psin(\bm{\dot{x}}_{2} - \bm{J}_{2}\bm{J}_{1}^\psin\bm{\dot{x}}_{1} ),
	\label{eq:null2}
\end{align}
where the first task is prioritized over the second one. This task priority formulation has a significant advantage over the standard formulation of \eqref{eq:stdnull}, because conflicting tasks can be handled more efficiently. This is mainly because \eqref{eq:null2} uses the nullspace of the augmented Jacobian matrix, whereas \eqref{eq:stdnull} only uses the nullspace of the Jacobian of the previous task. For more details, the reader can refer to Appendix \ref{app:nullspace}. 

This also can be extended to have $K$ tasks with the recursive formulation \cite{SicilianoSlotine91}
\begin{align}
	\bm{u}_{i+1} = \bm{u}_{i} + (\bm{J}_{i}\bm{P}_{i-1}^{A})^\psin(\bm{\dot{x}}_{i} - \bm{J}_{i}\bm{u}_i),
	\label{eq:null2K}
\end{align}
where $\bm{P}_{i}^A = \bm{I} - {\bm{J}_{i}^A}^\psin\bm{J}_{i}^A$ is the projection matrix onto the nullspace of the augmented Jacobian matrix $\bm{J}_{i}^A = [\bm{J}_{1}^\trsp \bm{J}_{2}^\trsp \ldots \bm{J}_{i}^\trsp]^\trsp$.

$\bm{P}_{i}^A$ can also be formulated recursively as 
\begin{align*}
	\bm{P}_{i}^A = \bm{P}_{i-1}^A - (\bm{J}_{i}\bm{P}_{i-1}^{A})^\psin(\bm{J}_{i}\bm{P}_{i-1}^{A}),
\end{align*}
with $\bm{P}_{0}^A = \bm{I}$, see \cite{Baerlocher98} for details.

In inverse kinematics, we can encounter kinematic singularities arising from the singularity of the Jacobian matrix which can result in high velocities that could potentially damage the robot. These singularities can be handled using a damped (or regularized) pseudoinverse computed as
\begin{align}
	\bm{J}^\psinr = \bm{J}^\trsp {(\bm{J} \bm{J}^\trsp + \lambda\bm{I}_{\bm{x}})}^{-1} = {(\bm{J}^\trsp \bm{J} + \lambda\bm{I}_{\bm{u}} )}^{-1} \bm{J}^\trsp,
	\label{eq:reg_ik}
\end{align}
where $\lambda$ is the damping factor. $\bm{I}_{\bm{x}}$ and $\bm{I}_{\bm{u}}$ are identity matrices with dimensions of $\bm{x}$ and $\bm{u}$, respectively. Using SVD, we can show the existence of the nullspace in the presence of damping (see Appendix \ref{app:nullspace}). Note that both standard and task-priority formulations can contain kinematic singularities. However, we can, in addition, have an algorithmic singularity in \eqref{eq:null2} because of the computation of $(\bm{J}_{2}\bm{N}_{1})^\psin$. Algorithmic singularities can also be handled using damped pseudoinverses \cite{Chiaverini97}.

In some cases, we may be more interested in using some specific joints $\bm{u}$ more than others, or putting more importance on accomplishing the task on a specific dimension of $\bm{x}$. By denoting $\bm{\tilde{J}} = \bm{U}_{\bm{x}}^\trsp\bm{J}\bm{U_q}^{\trsp^{-1}}$,  $\bm{\tilde{\dot{x}}} = \bm{U}_{\bm{x}}^\trsp\bm{\dot{x}}$ and $\bm{\tilde{\dot{q}}} = \bm{U_q}^\trsp \bm{\dot{q}}$ we obtain the weighted IK solution 
\begin{align}
	\bm{u}
	&=\bm{U_q}^{\trsp^{-1}}{\bm{\tilde{J}}}^\psin \bm{\tilde{\dot{x}}} + \bm{U_q}^{\trsp^{-1}}\bm{\tilde{N}}\bm{\tilde{\dot{q}}},
\end{align}
where $\bm{U_{x}}$ and $\bm{U_{q}}$ are decomposition (e.g., eigen or square root) matrices of weight matrices $\bm{W_{x}}$ and $\bm{W_{q}}$, respectively, such that $\bm{W_{x}} =\bm{U_{x}}\bm{U_{x}}^\trsp $ and $\bm{W_{q}} =\bm{U_{q}}\bm{U_{q}}^\trsp$. Computational details are given in Appendix \ref{app:nullspace}.

\section{Standard model predictive control}
\label{sec:MPC}

We consider the model predictive control (MPC) problem of estimating a controller $\bm{u}_t\!\in\!\mathbb{R}^{d}$ for a discrete linear dynamical system $\bm{x}_{t+1} = f(\bm{x}_t,\bm{u}_t)$, with state variable $\bm{x}_t\!\in\!\mathbb{R}^{DC}$, typically composed of position and velocity ($C=2$). The problem is formulated as the minimization of the cost
\begin{align}
	c
	&= {\big(\bm{\mu}_T\!-\!\bm{x}_T\big)}^\trsp
	\bm{Q}_T 
	\big(\bm{\mu}_T\!-\!\bm{x}_T\big) +
	\sum_{t=1}^{T-1} \Big({\big(\bm{\mu}_t\!-\!\bm{x}_t\big)}^\trsp
	\bm{Q}_t  
	\big(\bm{\mu}_t\!-\!\bm{x}_t\big) 
	\;+\;
	\bm{u}_t^\trsp \bm{R}_t\; \bm{u}_t \Big)
	\label{eq:ct}
\end{align}
subject to the linearization of $\bm{x}_{t+1} = f(\bm{x}_t,\bm{u}_t)$ expressed as
\begin{equation}
	\bm{x}_{t+1} = \bm{A}_t(\bm{x}_t,\bm{u}_t) \; \bm{x}_t + \bm{B}_t(\bm{x}_t,\bm{u}_t) \; \bm{u}_t.
	\label{eq:AB}
\end{equation}
Such problem can be solved by considering $\bm{x}\!=\!{\begin{bmatrix}\bm{x}_1^\trsp, \bm{x}_2^\trsp, \ldots, \bm{x}_T^\trsp \end{bmatrix}}^\trsp\!\in\!\mathbb{R}^{DCT}$ the evolution of the state variable, $\bm{u}\!=\!{\begin{bmatrix}\bm{u}_1^\trsp, \bm{u}_2^\trsp, \ldots, \bm{u}_{T-1}^\trsp \end{bmatrix}}^\trsp\!\in\!\mathbb{R}^{d(T-1)}$ the evolution of the control variable, $\bm{\mu}\!=\!{\begin{bmatrix}\bm{\mu}_1^\trsp, \bm{\mu}_2^\trsp, \ldots, \bm{\mu}_T^\trsp \end{bmatrix}}^\trsp\!\in\!\mathbb{R}^{DCT}$ the evolution of the tracking target, $\bm{Q}\!=\!\mathrm{blockdiag}(\bm{Q}_1,\bm{Q}_2,\ldots,\bm{Q}_T)\in\mathbb{R}^{DCT\times DCT}$ the evolution of the required tracking precision, and 
$\bm{R}\!=\!\mathrm{blockdiag}(\bm{R}_{1},\bm{R}_{2},\ldots,\bm{R}_{T-1})\in\mathbb{R}^{d(T-1)\times d(T-1)}$ the evolution of the cost on the control inputs. 

The constrained objective \eqref{eq:ct} then corresponds to
\begin{align}
	\bm{\hat{u}} &= \arg\min_{\bm{u}}  {\big(\bm{\mu}-\bm{x}\big)}^\trsp \bm{Q} \big(\bm{\mu}-\bm{x}\big)
	\;+\;
	\bm{u}^\trsp \!\bm{R} \bm{u} 
	\quad\mathrm{s.t.}\quad \bm{x}=\bm{S}_{\bm{x}}\bm{x}_1+\bm{S}_{\bm{u}}\bm{u}\nonumber\\
	&= {\big({\bm{S}_{\bm{u}}}^\trsp \bm{Q} \bm{S}_{\bm{u}} + \bm{R}\big)}^{-1}
	{\bm{S}_{\bm{u}}}^\trsp \bm{Q} 
	\big(\bm{\mu} - \bm{S}_{\bm{x}} \bm{x}_1 \big)\nonumber\\
	&= {(\bm{\tilde{J}}^\trsp \bm{\tilde{J}} + \lambda\bm{I}_{\bm{u}} )}^{-1} \bm{\tilde{J}}^\trsp\bm{\tilde{\dot{x}}}\nonumber\\
	&= \bm{\tilde{J}}^\psinr \, \bm{\tilde{\dot{x}}},
	\label{eq:stdMPC}
\end{align}
with transfer matrices $\bm{S}_{\bm{u}}$ and $\bm{S}_{\bm{x}}$ (see Appendix \ref{app:MPC} for details). $\bm{\tilde{J}} = \bm{U_{x}}^\trsp\bm{S}_{\bm{u}}$ and $\bm{\tilde{\dot{x}}} = \bm{U_{x}}^\trsp(\bm{\mu} - \bm{S}_{\bm{x}} \bm{x}) $ with decomposition $\bm{Q}=\bm{U_{x}}\bm{U_{x}}^\trsp$, and by assuming that $\bm{R} = \lambda\bm{I}_{\bm{u}} $.

\section{Model predictive control with nullspace structure}
\label{sec:MPC_null}

Recall that for LQR $\bm{\tilde{J}} = \bm{U_{x}}^\trsp\bm{S}_{\bm{u}}$, where $\bm{S}_{\bm{u}} \in \mathbb{R}^{DCT \times D(T-1)}$, $m = DCT$, $n = D(T-1)$ $r = \mathrm{rank}(\bm{S}_{\bm{u}}) = D(T-1)$. Sparse $\bm{U_x}$ are typically used in MPC problems, allowing the use of nullspace structures with a transfer matrix $\bm{S}_{\bm{u}}$ of reduced rank. A simple example is to reach a goal state $\bm{\mu}$ with a precision $\bm{Q}_T$ starting from an initial state $\bm{x}_0$. We can construct $\bm{Q}$ and $\bm{\mu}$ accordingly as $\bm{Q} = \mathrm{blockdiag}(\bm{0}, \bm{0}, \ldots, \bm{Q}_T)$ and $\bm{\mu}\!=\!{\begin{bmatrix}\bm{\mu}^\trsp, \bm{\mu}^\trsp, \ldots, \bm{\mu}^\trsp \end{bmatrix}}^\trsp$. We have then $\mathrm{rank}(\bm{\tilde{J}}) =\mathrm{rank}(\bm{Q}) = \mathrm{rank}(\bm{Q}_T)$ which is smaller than the original matrix. 

If we have sparse weight matrices, then the complete solution of \eqref{eq:stdMPC} can be rewritten as
\begin{align}
	\bm{\hat{u}} =
	{\bm{\tilde{J}}}^\psinr \, \bm{\tilde{\dot{x}}} + \bm{\tilde{N}} \, \bm{u}^\ty{(2)},
	\label{eq:MPC}
\end{align}
with LQR nullspace projection matrix $\bm{\tilde{N}}$ and any secondary control trajectory $\bm{u}^\ty{(2)}$. Instead of putting a secondary control trajectory, one can also replace it with the corresponding ${\bm{\tilde{J}}}_2^\psinr \bm{\tilde{\dot{x}}}_2$. 
In this paper, however, for the stability and robustness issues mentioned in \cite{Antonelli09}, we will use task-priority formulation \eqref{eq:null2K} as
\begin{align}
	\bm{\hat{u}}_{i+1} = \bm{\hat{u}}_{i} + \bm{\tilde{P}}_{i-1}^{A} \, (\bm{\tilde{J}}_{i}\bm{\tilde{P}}_{i-1}^{A})^\psinr \, (\bm{\dot{\tilde{x}}}_{i} - \bm{\tilde{J}}_{i}\bm{\hat{u}}_i),
	\label{eq:null_LQR}
\end{align}
where $\bm{\tilde{P}}_{i}^A = \bm{I} - {\bm{\tilde{J}}}_{i}^{A^\psinr}\bm{\tilde{J}}_{i}^A$ is the projection matrix onto the nullspace of the augmented task matrix (Jacobian matrix, in IK) $\bm{\tilde{J}}_{i}^A = [\bm{\tilde{J}}_{1}^\trsp \bm{\tilde{J}}_{2}^\trsp \ldots \bm{\tilde{J}}_{i}^\trsp]^\trsp$.
$\bm{\tilde{P}}_{i}^A$ can also be formulated recursively as 
\begin{align}
	\bm{\tilde{P}}_{i}^A = \bm{\tilde{P}}_{i-1}^A - \bm{\tilde{P}}_{i-1}^{A} \, (\bm{J}_{i}\bm{\tilde{P}}_{i-1}^{A})^\psin \, (\bm{J}_{i}\bm{\tilde{P}}_{i-1}^{A}),
\end{align}
with $\bm{\tilde{P}}_{0}^A = \bm{I}$. Note that this is not exactly \eqref{eq:null2K}, since with regularization, we lose the properties of idempotent matrices (explained in Appendix \ref{app:nullspace}), and we do not have $\bm{\tilde{P}}_{i-1}^{A}(\bm{J}_{i}\bm{\tilde{P}}_{i-1}^{A})^\psin = (\bm{J}_{i}\bm{\tilde{P}}_{i-1}^{A})^\psin$ anymore.

Nullspace structure in MPC allows us to exploit the redundancy in space-time. At each time step, we can have tasks that require different precisions redundant in space dimensions, with different priorities. In the same way, we can have tasks that are redundant in time dimensions for each space dimension.

If we want to have $p$ tasks $\{\bm{\tilde{J}}_{i}\}_{i=1}^p$ with the same priorities, we can use an augmented task matrix $\bm{\tilde{J}}_{i}^A = [\bm{\tilde{J}}_{i}^\trsp \bm{\tilde{J}}_{i+1}^\trsp \ldots \bm{\tilde{J}}_{p}^\trsp]^\trsp$ to find an optimal solution that compromises between these $p$ tasks, without any priorities. This solution is also called \textit{fusion} and can be represented by a product of Gaussians \cite{Calinon16HFR}. 


\section{Experiments}
\label{sec:Experiments}

\subsection{Proof-of-concept examples}

\begin{figure}[ht!]
	\centering
	\includegraphics[width=.9\columnwidth]{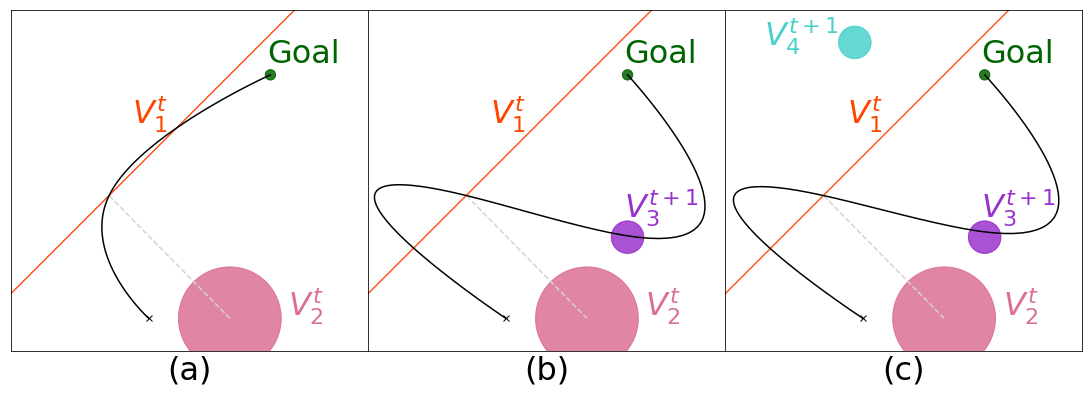}
	\caption{
LQR with initially at the position shown by the cross and (\emph{a}) 2 intermediary tasks at time step $t$ with hierarchies $\textit{V}_{1}^t > \textit{V}_{2}^t$ and the final task $\textit{Goal}$, 
(\emph{b}) 2 intermediary tasks at time step $t$, 1 intermediary task at time step $t+1$ with hierarchies $\textit{V}_{1}^t > \textit{V}_{2}^t > \textit{V}_{3}^{t+1}$, and the final task $\textit{Goal}$, 
(\emph{c}) 2 intermediary tasks at time step $t$, 2 intermediary tasks at time step $t+1$ with hierarchies $\textit{V}_{1}^t > \textit{V}_{2}^t > \textit{V}_{3}^{t+1} > \textit{V}_{4}^{t+1} $, and the final task $\textit{Goal}$. Tasks are represented with Gaussian ellipsoids, while the resulting trajectory is represented by black lines. Grey dashed line represents the shortest line between the mean of the Gaussian ellipsoid of $ \textit{V}_{2}^t$ and the line ellipsoid (thin Gaussian ellipsoid) $ \textit{V}_{1}^t$.
	}
	\label{fig:LQR_ex}
\end{figure}

Our first example consists in reaching a goal position \textit{Goal}, while passing through viapoints with different precisions and different hierarchies, see Figure \ref{fig:LQR_ex}. The primary viapoint represented by $\textit{V}_{1}^t$ is to be on the line (depicted by thin ellipsoid) at time step $t$ (used as half-time of the execution for the plot). This task creates redundancy in space-time because according to the upcoming secondary tasks and previous actions taken, the position on the line at time step $t$ can change. The secondary viapoint represented by $\textit{V}_{2}^t$ has an isotropic precision with no redundancy in any space dimension, meaning that at time step $t$, the objective of the secondary task is to be at the center of the pink circle. Figure \ref{fig:LQR_ex}(a) shows an LQR execution with a unit mass double integrator starting from the initial position (shown by the cross), with 2 viapoint tasks at time step $t$ with hierarchies $\textit{V}_{1}^t > \textit{V}_{2}^t$ and the final task $\textit{Goal}$. At time step $t$, the algorithm finds the best compromise between accomplishing both tasks, taking into account also the priorities. Such best compromise can be found intuitively as the intersection point between the thin ellipse and projection of the center of the pink circle onto the thin ellipse, hence the dashed gray line.

Figure \ref{fig:LQR_ex}(b) shows the trajectory of the same agent, with the addition of a tertiary task $\textit{V}_{3}^{t+1}$ to be achieved at time step $t+1$ and whose variance is represented by a purple circle. Since this position can be reached without disturbing the primary and secondary tasks, the trajectory changes so as to accomplish all three tasks. Figure \ref{fig:LQR_ex}(c) shows that the addition of another quaternary task $\textit{V}_{4}^{t+1}$ to be achieved at time step $t+1$, represented by a turquoise circle, does not change the trajectory shown in Figure \ref{fig:LQR_ex}(b)  because this task is in conflict with the tertiary task and is thus neglected due to its priority. 

\begin{figure}[ht!]
	\centering
	\includegraphics[width=.6\columnwidth]{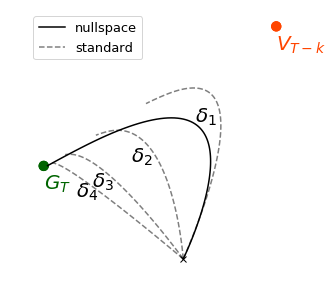}
	\caption{
Comparison between nullspace (solid black line) and standard formulation (dashed lines) with scaling $\delta_1$, $\delta_2$, $\delta_3$ and $\delta_4$ are $1$, $0.1$, $0.01$ and $0.001$ respectively, using a point mass object with double integrator dynamics. The agent has to go to its goal point $G_T$ as its primary task, at the final time step $T$, with variance shown by a green circle, and pass through a viapoint $V_{T-k}$, as a secondary task, at the time step $T-k$, with same variance shown by orange circle. When the secondary task is far away from the primary task in space and time, nullspace planning tries to pass through $V_{T-k}$, and is still successful to accomplish the primary task. On the other hand, no matter scaling of the cost, standard control cannot have the same performance.
	}
	\label{fig:fusionVSnull}
\end{figure}

Figure \ref{fig:fusionVSnull} shows an example where the nullspace formulation has an advantage over the naive attempt of minimizing both tasks' errors at the same, using an hyperparameter $\delta$ to set importance weights between the tasks, where $\delta = 0$ achieves only the primary task and a large $\delta$ achieves only the secondary task. In this example, we want to reach the primary point $G_T$ at the final time step $T$ with a given variance represented by green circle as the primary task. We also want to pass through the viapoint $V_{T-k}$, $k$ time steps before the final time step $T$, with a given variance represented by orange circle as the secondary task. With real-world robots, we have restrictions on the norm of the control commands, hence achieving these both tasks would become impossible if they are too far away in space dimensions but very close in time dimensions. In this Figure, any naive attempts to accomplish the tasks with a hierarchy imposed by $\delta$ fails, except for very small values, which are not interesting, because we know intuitively that we can still do much better by trying to achieve the secondary task than achieving only the primary task. 

\subsection{Robot Simulation}

\begin{figure}
	\centering
	\subfloat[Without nullspace]{{\includegraphics[width=.46\columnwidth]{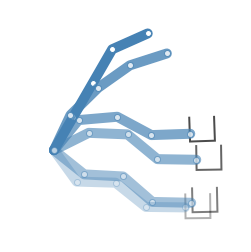} }}
	\qquad
	\subfloat[With nullspace]{{\includegraphics[width=.46\columnwidth]{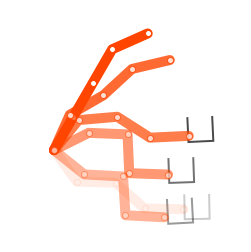} }}
	\caption{
A robotic application of nullspace structure within LQR.
	}
	\label{fig:robot_sim}
\end{figure}

The setup consists of two robotic agents, represented by point mass agents shown in Figure \ref{fig:robot_sim} (black and red). They have to reach a goal position as a primary task while meeting with each other at the halfway of their movement as a secondary task. One can think of many real life applications that can be represented by this example: a bimanual robot passing an object from one hand to the other and then using both hands individually, or two mobile robots that have to exchange some products in a factory before moving to their respective location. These applications require the robots to have multiple layers of workload, such as safety, main mission completion, social navigation, etc. Considering that these two agents are perturbed during their execution to, for example, avoid some obstacles, then we expect these robots to perform their main mission and do the secondary tasks in the nullspace. In this figure, we see that any perturbation that is not in conflict between the main task of reaching the goal can be performed using the nullspace structure by autonomously modifying the meeting position.
 
\begin{figure}[ht!]
	\centering
	\includegraphics[width=1\columnwidth]{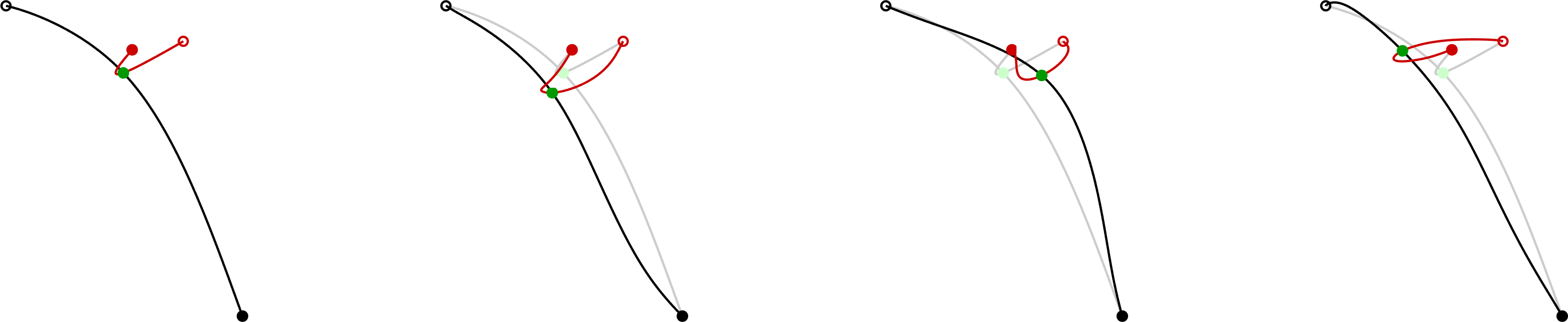}
	\caption{
Two agents (black and red) have a primary task of reaching their goal positions at time step $T$ and a secondary goal to meet at some position, at time step $T/2$. If obstacles perturb their executions, we observe the nullspace effect with a shift of the meeting position.
	}
	\label{fig:LQR_ex2}
\end{figure}

We can think of another robotic application where we can exploit the redundancies in time-space dimensions of the robot. In Fig.~\ref{fig:robot_sim}, we have a 4 DoF robot controlled by acceleration commands in $T$ time steps. The state, consisting of 4-dimensional joint positions and 4-dimensional joint velocities, is 8-dimensional. Therefore we have a total of $8T$ DoF along the trajectory of the robot. The robot has to pick up a cup at time step $T/2$ and place it at another location at time step $T$. We assume that picking up the cup and placing it both spend 8 DoF, which makes a total of 16 DoF used. The resulting trajectory is shown in Figure \ref{fig:robot_sim}(a). Then, we impose a secondary objective, after picking up the cup, to hold a 90 degree angle for its last 2 joints. This is helpful for the robot to hold the cup with a better manipulability, as a human would do. Note that the secondary objective uses only 2 DoF at each time step, with a total of $2\times(T/2)$ DoF used. The resulting trajectory of Figure \ref{fig:robot_sim}\emph{(b)} shows that the robot is able to keep 90 degree angles only when it is not in conflict with the tasks of picking up the cup and placing it. 

\clearpage
\appendix
\section*{Appendices}

\section{Computation of nullspace control}
\label{app:nullspace}

The nullspace control problem can be formulated as the constrained objective
\begin{align}
	\min_{\bm{u}} {\big\|\bm{u} - \bm{\dot{q}} \big\|}^2 
	\quad\mathrm{s.t.}\quad 
	\bm{J}\bm{u}=\bm{\dot{x}},
	\label{eq:chiaverini}
\end{align}
which can be solved through Lagrange multipliers, by defining the objective
\begin{align*}
	\min_{\bm{u}} {\big\|\bm{u} - \bm{\dot{q}} \big\|}^2 
	+\bm{\lambda}^\trsp (\bm{J}\bm{u}-\bm{\dot{x}}).
\end{align*}
Differentiating with respect to $\bm{u}$ and $\bm{\lambda}$ and equating to zero gives
\begin{align}
	\bm{u} - \bm{\dot{q}} + \bm{J}^\trsp \bm{\lambda} &= \bm{0}, \label{eq:udiff}\\
	\bm{J} \bm{u} &= \bm{\dot{x}},
\end{align}
whose combination results in 
\begin{align}
	\nonumber
	&\bm{J} \bm{\dot{q}} - \bm{J} \bm{J}^\trsp \bm{\lambda} = \bm{\dot{x}}\\
	\iff\quad
	&\bm{\lambda} = {(\bm{J} \bm{J}^\trsp)}^{-1} (\bm{J} \bm{\dot{q}} - \bm{\dot{x}}). \label{eq:lambda}
\end{align}
By reintroducing \eqref{eq:lambda} into \eqref{eq:udiff}, we then get
\begin{align}
	\nonumber
	&\bm{u} - \bm{\dot{q}} + \underbrace{\bm{J}^\trsp {(\bm{J} \bm{J}^\trsp)}^{-1}}_{\bm{J}^\psin} (\bm{J} \bm{\dot{q}} - \bm{\dot{x}}) = \bm{0} \\
	\iff\quad 
	&\bm{u} = \bm{J}^\psin \bm{\dot{x}} +  \underbrace{(\bm{I} - \bm{J}^\psin \bm{J})}_{\bm{N}} \bm{\dot{q}}.
\end{align}

\subsection{Pseudoinverse computation}

We consider the SVD decomposition
\begin{align}
	\bm{J} = 
	\begin{bmatrix}
	\bm{U}_1& \bm{U}_2
	\end{bmatrix}
	\begin{bmatrix}
	\bm{\Sigma}_{1}&\bm{0} \\ 
	\bm{0}&\bm{0} 
	\end{bmatrix}
	\begin{bmatrix}
	\bm{V}_1^\trsp\\ \bm{V}_2^\trsp
	\end{bmatrix},
\end{align} 
for any $\bm{J} \in \mathbb{R}^{m \times n}$. Here $\bm{U}_{1} \in\mathbb{R}^{m \times r} $, $\bm{U}_{2} \in\mathbb{R}^{m \times m-r} $, $\bm{V}_{1}^\trsp \in\mathbb{R}^{r \times n} $, $\bm{V}_{2}^\trsp \in\mathbb{R}^{n-r \times n} $ and $\bm{\Sigma}_{1}^\trsp \in\mathbb{R}^{r \times r} $, where $r = \text{rank}(\bm{J})$ and $\bm{\Sigma}_{1}$ is the diagonal matrix of non-zero singular values of $\bm{J}$. Then we can define the pseudoinverse and the nullspace matrix as
\begin{align}
	\bm{J}^\psin &= \bm{V}_{1}\bm{\Sigma}_{1}^{-1} \bm{U}_{1}^\trsp,\\
	\bm{N} &= \bm{V}_{2}\bm{V}_{2}^\trsp.
\end{align}
Using these notations, one can notice that the only time the nullspace matrix does not exist, is when $r = n$ since $\bm{V}_{2}$ does not exist (see the dimensions). 

A regularized pseudoinverse matrix and the corresponding nullspace matrix can be written as
\begin{align}
	\bm{J}^\psinr &= \bm{V}_1\bm{\Sigma}_1^{\psinr} \bm{U}_1^\trsp,\label{eq:psinr_svd}\\
	\bm{N}&= \bm{V}_{2}\bm{V}_{2}^\trsp + \bm{V}_{1}(\bm{I} -\bm{\Sigma}_{1}^{\psinr}\bm{\Sigma}_1 )\bm{V}_{1}^\trsp,
	\label{eq:nullr_svd}
\end{align}
where $\bm{\Sigma}_1^{\psinr} = \bm{\Sigma}_1(\bm{\Sigma}_1^2 + \lambda \bm{I})^{-1}$. Notice that in the case of $r = n$, the nullspace (that did not exist before) has now some value because even though $\bm{V}_{2}$ does not exist, the second term in \eqref{eq:nullr_svd} is not null and can be denoted as an error term in the calculation of nullspace matrix when we use regularization. 

\subsection{Extension to a weighted problem}

We would like to find a solution to 
\begin{align}
	\min_{\bm{u}}{\frac{1}{2}\big\|\bm{u} - \bm{\dot{q}} \big\|}_{\bm{W}_{\bm{q}}}^2,
	\label{eq:second_task}
\end{align}
among all the solutions of
\begin{align}
	\min_{\bm{u}}{\frac{1}{2}\big\|\bm{J}\bm{u} - \bm{\dot{x}} \big\|}_{\bm{W}_{\bm{x}}}^2.
	\label{eq:first_task}
\end{align}
In other words, in all the solution set that satisfy the first task, found by minimizing \eqref{eq:first_task}, we would like to get the ones that satisfy also \eqref{eq:second_task}. This is the interpretation of \eqref{eq:chiaverini}, without $\bm{W_{x}}$ and $\bm{W_{q}}$.
We can transform these weighted problems of \eqref{eq:first_task} and \eqref{eq:second_task} into a "nonweighted" one as in \eqref{eq:chiaverini} by setting
\begin{align}
	\bm{\tilde{J}} = \bm{U}_{\bm{x}}^\trsp\bm{J}\bm{U_q}^{\trsp^{-1}} \text{,} \quad \bm{\tilde{\dot{x}}} = \bm{U}_{\bm{x}}^\trsp\bm{\dot{x}}  \quad \text{,} \quad
	\bm{\tilde{\dot{q}}} = \bm{U_{q}}^\trsp\bm{\dot{q}} \quad \text{,} \quad
	\bm{\tilde{u}} = \bm{U_{q}}^\trsp\bm{u},
\end{align}
and obtaining
\begin{align}
	\min_{\bm{\tilde{u}}} {\frac{1}{2}\big\|\bm{\tilde{u}} - \bm{\tilde{\dot{q}}} \big\|}^2 
	\quad\mathrm{s.t.}\quad 
	\bm{\tilde{J}}\bm{\tilde{u}}=\bm{\tilde{\dot{x}}}.
	\label{eq:chiaverini_weighted}
\end{align}
Then, the solution of \eqref{eq:chiaverini_weighted} should be transformed back to the original space. Equivalence of weighted formulations to transformed unweighted formulations can easily be verified by
\begin{align*}
	{\big\|\bm{u} - \bm{\dot{q}} \big\|}_{\bm{W}_{\bm{q}}}^2 
	&= (\bm{u} - \bm{\dot{q}})^\trsp \bm{W_q} (\bm{u} - \bm{\dot{q}}) \\
	&= (\bm{u} - \bm{\dot{q}})^\trsp \bm{U_q}\bm{U_q}^\trsp (\bm{u} - \bm{\dot{q}}) \\
	&= (\bm{U_q}^\trsp\bm{u} - \bm{U_q}^\trsp\bm{\dot{q}})^\trsp (\bm{U_q}^\trsp\bm{u} - \bm{U_q}^\trsp\bm{\dot{q}}) \\
	&= (\bm{\tilde{u}} - \bm{\tilde{\dot{q}}})^\trsp (\bm{\tilde{u}} - \bm{\tilde{\dot{q}}})\\
	&= {\big\|\bm{\tilde{u}} - \bm{\tilde{\dot{q}}} \big\|}^2,
\end{align*}
and
\begin{align*}
	{\big\|\bm{J}\bm{u} - \bm{\dot{x}} \big\|}_{\bm{W}_{\bm{x}}}^2 
	&= (\bm{J}\bm{u} - \bm{\dot{x}})^\trsp \bm{W_x} (\bm{J}\bm{u} - \bm{\dot{x}}) \\
	&= (\bm{J}\bm{u} - \bm{\dot{x}})^\trsp \bm{U_x}\bm{U_x}^\trsp (\bm{J}\bm{u} - \bm{\dot{x}}) \\
	&= (\bm{U_x}^\trsp\bm{J}\bm{u} - \bm{U_x}^\trsp\bm{\dot{x}})^\trsp (\bm{U_x}^\trsp\bm{J}\bm{u} - \bm{U_x}^\trsp\bm{\dot{x}}) \\
	&= (\bm{U_x}^\trsp\bm{J}\bm{U_q}^{\trsp^{-1}}\bm{U_q}^\trsp\bm{u} - \bm{U_x}^\trsp\bm{\dot{x}})^\trsp (\bm{U_x}^\trsp\bm{J}\bm{U_q}^{\trsp^{-1}}\bm{U_q}^\trsp\bm{u} - \bm{U_x}^\trsp\bm{\dot{x}}) \\
	&= (\bm{\tilde{J}}\bm{\tilde{u}} - \bm{\tilde{\dot{x}}})^\trsp (\bm{\tilde{J}}\bm{\tilde{u}} - \bm{\tilde{\dot{x}}})\\
	&= {\big\|\bm{\tilde{J}}\bm{\tilde{u}} - \bm{\tilde{\dot{x}}} \big\|}^2.
\end{align*}
The weights $\bm{W_{q}}$ on the joint space (i.e., on minimization variable space) affect all the tasks that we want to accomplish. Therefore, we can only have one $\bm{W_{q}}$. Alternatively, we can have different $\bm{W}_{\bm{x}_{i}}$ for each task $i$. When we have the $i^\text{th}$ task described in joint space, namely $\bm{\dot{q}}_{i}$, then the appropriate transformations are $\bm{\tilde{\dot{q}}}_{i} = \bm{U}_{\bm{q}}^\trsp\bm{\dot{q}}_{i}$ and $\bm{\tilde{u}} = \bm{U}_{\bm{q}}^\trsp\bm{u}$. When we have the $i^\text{th}$ task described in task space, namely $\bm{J}_{i}^\psin\bm{\dot{x}}_{i}$, then the appropriate transformations are $\bm{\tilde{J}}_{i} = \bm{U}_{\bm{x}_{i}}^\trsp\bm{J}_{i}\bm{U_q}^{\trsp^{-1}}$ and $\bm{\tilde{\dot{x}}}_{i} = \bm{U}_{\bm{x}_{i}}^\trsp\bm{\dot{x}}_{i}$. After solving the nullspace weighted control problem with appropriate transformations, one should transform back into the original space of minimization variable.

\subsection{Task priority formulation}

A task-priority formulation \cite{Hanafusa32} describes a solution for the nullspace control of inverse kinematics problem as
\begin{align}
	\bm{u} = \bm{J}_{1}^\psin\bm{\dot{x}}_{1} + (\bm{J}_{2}\bm{N}_{1})^\psin(\bm{\dot{x}}_{2} - \bm{J}_{2}\bm{J}_{1}^\psin\bm{\dot{x}}_{1} ),
	\label{eq:null2_}
\end{align}
where the first task is prioritized over the second one. \\

\noindent \textbf{Proof:}
Consider the solution of the hierarchically the first task as
\begin{align}
	\bm{u} = \bm{J}_{1}^\psin\bm{\dot{x}}_{1} + \bm{N}_{1}\bm{y},
	\label{eq:first_task_}
\end{align}
where $\bm{y}$ is arbitrary. We would like this solution to satisfy the secondary task $\bm{J}_{2}\bm{u} = \bm{\dot{x}}_{2}$ as well. By plugging \eqref{eq:first_task_}, we obtain
\begin{align}
	\bm{J}_{2}( \bm{J}_{1}^\psin\bm{\dot{x}}_{1} + \bm{N}_{1}\bm{y}) =\bm{\dot{x}}_{2},
\end{align}
and find
\begin{align}
	\bm{y} = (\bm{J}_{2}\bm{N}_{1})^\psin(\bm{\dot{x}}_{2} -\bm{J}_{2}\bm{J}_{1}^\psin\bm{\dot{x}}_{1} )
	\label{eq:y_sol}.
\end{align}
We can substitute \eqref{eq:y_sol} back into \eqref{eq:first_task_} to obtain \eqref{eq:null2_} as
\begin{align}
	\bm{u} &= \bm{J}_{1}^\psin\bm{\dot{x}}_{1} + \bm{N}_{1}(\bm{J}_{2}\bm{N}_{1})^\psin(\bm{\dot{x}}_{2} -\bm{J}_{2}\bm{J}_{1}^\psin\bm{\dot{x}}_{1} )\nonumber \\	
	&=  \bm{J}_{1}^\psin\bm{\dot{x}}_{1} + (\bm{J}_{2}\bm{N}_{1})^\psin(\bm{\dot{x}}_{2} - \bm{J}_{2}\bm{J}_{1}^\psin\bm{\dot{x}}_{1} ),
\end{align}
where we use the property of $\bm{N}_{1}$ being idempotent and hermitian, i.e., $\bm{N}_{1}^2 = \bm{N}_{1}$ and $\bm{N}_{1}^\trsp = \bm{N}_{1}$, so that 
$\bm{N}_{1}(\bm{J}_{2}\bm{N}_{1})^\psin =(\bm{J}_{2}\bm{N}_{1})^\psin$.

\section{Computation of nullspace control as PoG}
\label{app:nullspacePoG}

One of the advantages of standard nullspace control over a task priority formulation is that we can formulate the former as a product of Gaussians. By denoting the pseudoinverse and nullspace projection matrices as
\begin{align*}
	\bm{J}^\psin &= \bm{J}^\trsp {(\bm{J} \bm{J}^\trsp)}^{-1},\\
	\bm{N} &= \bm{I} - \bm{J}^\psin \bm{J}, 
\end{align*}
and by considering two Gaussians $\mathcal{N}(\bm{\mu}_1,\bm{\Gamma}_1^{-1})$ and $\mathcal{N}(\bm{\mu}_2,\bm{\Gamma}_2^{-1})$ with parameters\\
\begin{minipage}{.48\columnwidth}
	\begin{align*}
	\bm{\mu}_1 &= \bm{J}^\psin \bm{\dot{x}},\\
	\bm{\Gamma}_1 &= \bm{J}^\psin \bm{J},\\
	\end{align*}
\end{minipage}
\begin{minipage}{.48\columnwidth}
	\begin{align*}
	\bm{\mu}_2 &= \bm{\dot{q}},\\
	\bm{\Gamma}_2 &= \bm{N},\\
	\end{align*}
\end{minipage}\\
we can write the product of these two Gaussians as $\mathcal{N}(\bm{\hat{\mu}},\bm{\hat{\Sigma}})$, with parameters
\begin{align*}
	\bm{\hat{\mu}} &= {(\bm{\Gamma}_1+\bm{\Gamma}_2)}^{-1} \left(\bm{\Gamma}_1 \bm{\mu}_1 + \bm{\Gamma}_2 \bm{\mu}_2\right) \\
	&= \underbrace{{(\bm{J}^\psin \bm{J}+\bm{I} - \bm{J}^\psin \bm{J})}^{-1}}_{\bm{I}} \left(\bm{J}^\psin \bm{J} \bm{J}^\psin \bm{\dot{x}} + \bm{N} \bm{\dot{q}}\right) \\
	&= \bm{J}^\psin \underbrace{\bm{J} \bm{J}^\trsp {(\bm{J} \bm{J}^\trsp)}^{-1}}_{\bm{I}} \bm{\dot{x}} + \bm{N} \bm{\dot{q}} \\
	&= \bm{J}^\psin \bm{\dot{x}} + \bm{N} \bm{\dot{q}}, \\
	\bm{\hat{\Sigma}} &= {(\bm{\Gamma}_1+\bm{\Gamma}_2)}^{-1} \\
	&= {(\bm{J}^\psin \bm{J} + \bm{I} - \bm{J}^\psin \bm{J})}^{-1} \\
	&= \bm{I},
\end{align*}
which is here found by solving the quadratic optimization problem
\begin{align*}
	\bm{\hat{\mu}} &= \arg\min_{\bm{u}} {\big\|\bm{u} - \bm{J}^\psin \bm{\dot{x}}\big\|}_{\bm{J}^\psin \bm{J}}^2 + {\big\|\bm{u} - \bm{\dot{q}}\big\|}_{\bm{N}}^2 \\
	&= \bm{J}^\psin \bm{\dot{x}} + \bm{N} \bm{\dot{q}},
\end{align*}
corresponding to the standard nullspace control formulation.

For three task case, we should first note that $\bm{P} = \bm{J}^\psin\bm{J}$ and $\bm{N} = \bm{I}-\bm{P}$ are symmetric idempotent matrices with eigenvalues 0 or 1, therefore positive semi-definite, since they satisfy the property $\bm{P}^2 = \bm{P} = \bm{P}^\trsp $. Therefore,using these properties and considering the Gaussian parameters as\\
\begin{minipage}{.31\columnwidth}
	\begin{align*}
	\bm{\mu}_1 &= \bm{J}_{1}^\psin \bm{\dot{x}}_{1},\\
	\bm{\Gamma}_1 &= \bm{J}_{1}^\psin \bm{J}_{1},\\
	\end{align*}
\end{minipage}
\begin{minipage}{.31\columnwidth}
	\begin{align*}
	\bm{\mu}_2 &= (\bm{J}_{2}\bm{N}_{1})^\psin\bm{J}_{2}^\psin \bm{\dot{x}}_{2},\\
	\bm{\Gamma}_2 &= \bm{N}_{1}\bm{J}_{2}^\psin \bm{J}_{2}\bm{N}_{1},\\
	\end{align*}
\end{minipage}
\begin{minipage}{.31\columnwidth}
	\begin{align*}
	\bm{\mu}_3 &= (\bm{N}_{2}\bm{N}_{1})^\psin\bm{J}_{3}^\psin \bm{\dot{x}}_{3},\\
	\bm{\Gamma}_3 &= \bm{N}_{1}\bm{N}_{2}\bm{N}_{1},\\
	\end{align*}
\end{minipage}\\
we can then write the product of these three Gaussians as $\mathcal{N}(\bm{\hat{\mu}},\bm{\hat{\Sigma}})$, with parameters
\begin{align}
	\bm{\hat{\Sigma}} &=
	{(\bm{\Gamma}_1+\bm{\Gamma}_2 + \bm{\Gamma}_3 )}^{-1}
	\nonumber \\
	&=(\bm{J}_{1}^\psin \bm{J}_{1} + \bm{N}_{1}\bm{J}_{2}^\psin \bm{J}_{2}\bm{N}_{1} +\bm{N}_{1}\bm{N}_{2}\bm{N}_{1})^{-1}
	\nonumber \\
	&=\big(\bm{I}-\bm{N}_{1} + \bm{N}_{1}(\bm{I}-\bm{N}_{2})\bm{N}_{1} + \bm{N}_{1}\bm{N}_{2}\bm{N}_{1} \big)^{-1}
	\nonumber \\
	&= (\bm{I}-\bm{N}_{1} + \bm{N}_{1}\bm{N}_{1} - \bm{N}_{1}\bm{N}_{2}\bm{N}_{1} + \bm{N}_{1}\bm{N}_{2}\bm{N}_{1} )^{-1}
	\nonumber \\
	&= \bm{I},
	\\ 
	\bm{\hat{\mu}} &= {(\bm{\Gamma}_1+\bm{\Gamma}_2 + \bm{\Gamma}_3 )}^{-1} \left(\bm{\Gamma}_1 \bm{\mu}_1 + \bm{\Gamma}_2 \bm{\mu}_2+\bm{\Gamma}_3 \bm{\mu}_3\right) 
	\nonumber \\
	&= \bm{J}_{1}^\psin \bm{J}_{1}\bm{J}_{1}^\psin \bm{\dot{x}}_{1} + \bm{N}_{1}\bm{J}_{2}^\psin \bm{J}_{2}\bm{N}_{1}(\bm{J}_{2}\bm{N}_{1})^\psin\bm{J}_{2}^\psin \bm{\dot{x}}_{2}
	\nonumber \\
	&\quad   +\bm{N}_{1}\bm{N}_{2}\bm{N}_{1}(\bm{N}_{2}\bm{N}_{1})^\psin\bm{J}_{3}^\psin \bm{\dot{x}}_{3} 
	\nonumber \\
	&= \bm{J}_{1}^\psin\bm{\dot{x}}_{1} + \bm{N}_{1}\bm{J}_{2}^\psin\bm{\dot{x}}_{2} + \bm{N}_{1}\bm{N}_{2}\bm{J}_{3}^\psin \bm{\dot{x}}_{3}.
\end{align}

If more than two tasks of different priorities are described in task space, with Jacobians $\left \{\bm{J}_k\right \}_{k=1}^{K}$, the quadratic optimization problem becomes
\begin{align}
	\bm{\hat{\mu}} &=
	\arg\min_{\bm{u}} 
	{\big\|\bm{u} - \bm{J}_1^\psin \bm{\dot{x}}_1 \big\|}_{\bm{J}_1^\psin \bm{J}_1}^2 
	\nonumber \\
	&\quad
	+\sum_{k=2}^{K-1}{\big\|\bm{u} - (\bm{J}_k^\psin{\bm{N}^{(k)}}^\trsp) \bm{\dot{x}}_k  \big\|}_{{\bm{N}^{(k)}}{\bm{J}_k}^\psin \bm{J}_k{\bm{N}^{(k)}}^\trsp}^2 
	\nonumber \\
	&\quad
	+{\big\|\bm{u} - {{\bm{N}^{(K)}}^\trsp}^\psin {\bm{J}_K}^\psin \bm{\dot{x}}_K \big\|}_{{\bm{N}^{(K)}}^\trsp\bm{N}^{(K)}}^2,
	\label{eq:nullspace3}
\end{align}
where ${\bm{N}^{(k)}} = \prod_{j=1}^{k-1}\bm{N}_{j}$, the nullspace projection matrix $\bm{N}_{j} = (\bm{I}-\bm{J}_{j-1}^\psin \bm{J}_{j-1})$ and the number of tasks $K>2$.

\subsection{Extension to a weighted formulation with PoG}

Nullspace control can be extended to weighted nullspace control using
\begin{equation}
	\bm{\hat{\mu}} = \arg\min_{\bm{\tilde{u}}} 
	{\big\| \bm{u} - \bm{U_{q}}\bm{\tilde{J}}^\psin \bm{\tilde{\dot{x}}} \big\|}_{\bm{U_{q}}^{-1}{\bm{\tilde{J}}}^\psin \bm{\tilde{J}}\bm{U_{q}}^{-1}}^2 
	+{\big\| \bm{u} -\bm{\dot{q}} \big\|}_{\bm{U_{q}}^{-1}(\bm{I} - \bm{\tilde{J}}^\psin \bm{\tilde{J}}){\bm{U_{q}}}^{-1}}^2,
	\label{eq:weightednullspace}
\end{equation}
where $\bm{\tilde{J}} = \bm{U}_{\bm{x}}^\trsp\bm{J}\,{\bm{U}_{\bm{q}}^\trsp}^{-1}$ and $\bm{\tilde{\dot{x}}} = \bm{U}_{\bm{x}}^\trsp\bm{\dot{x}}$.

\section{Linear system evolution in MPC}
\label{app:MPC}

The expression $\bm{x}=\bm{S}_{\bm{x}}\bm{x}_1+\bm{S}_{\bm{u}}\bm{u}$ in \eqref{eq:stdMPC} can be found by expressing all future states $\bm{x}_t$ as an explicit function of the state $\bm{x}_1$. By writing
\begin{align*}
	\bm{x}_{2} &= \bm{A}_1 \bm{x}_1 + \bm{B}_1 \bm{u}_1 ,\\
	\bm{x}_{3} &= \bm{A}_2 \bm{x}_2 + \bm{B}_2 \bm{u}_2 = \bm{A}_2 (\bm{A}_1 \bm{x}_1 + \bm{B}_1 \bm{u}_1) + \bm{B}_2 \bm{u}_2 ,\\[-2mm]
	&\vdots\\[-2mm]
	\bm{x}_{T} &= \prod_{t=1}^{T-1} \bm{A}_{T-t} \bm{x}_1 + \prod_{t=1}^{T-2} \bm{A}_{T-t} \bm{B}_1 \bm{u}_1 
	 + \prod_{t=1}^{T-3} \bm{A}_{T-t} \bm{B}_2 \bm{u}_2 + \cdots + \bm{B}_{T-1} \bm{u}_{T-1},
\end{align*}
in a matrix form, we get an expression of the form $\bm{x}=\bm{S}_{\bm{x}}\bm{x}_1+\bm{S}_{\bm{u}}\bm{u}$, with
\begin{multline}
	\underbrace{
 	\begin{bmatrix}
 	\bm{x}_1\\
 	\bm{x}_2\\
 	\bm{x}_3\\
 	\vdots\\
 	\bm{x}_T
 	\end{bmatrix}}_{\bm{x}}
 	=
 	\underbrace{
 	\begin{bmatrix}
 	\bm{I}\\
 	\bm{A}_1\\
 	\bm{A}_2\bm{A}_1\\
 	\vdots\\
 	\prod_{t=1}^{T-1} \bm{A}_{T-t}
 	\end{bmatrix}}_{\bm{S}_{\bm{x}}}
 	\bm{x}_1
	+
 	\underbrace{
 	\begin{bmatrix}
 	\bm{0} & \bm{0} & \cdots & \bm{0} \\
 	\bm{B}_1 & \bm{0} & \cdots & \bm{0} \\
	\bm{A}_2\bm{B}_1 & \bm{B}_2 & \cdots & \bm{0} \\
	\vdots & \vdots & \ddots & \vdots \\
	\prod_{t=1}^{T-2} \bm{A}_{T-t} \bm{B}_1 & \prod_{t=1}^{T-3} \bm{A}_{T-t} \bm{B}_2 & \cdots & \bm{B}_{T-1}
	\end{bmatrix}}_{\bm{S}_{\bm{u}}}
	\underbrace{
	\begin{bmatrix}
	\bm{u}_1\\
 	\bm{u}_2\\
 	\vdots\\
 	\bm{u}_{T\!-\!1}
 	\end{bmatrix}}_{\bm{u}}
 	,
 	\label{eq:dynSysBatch}
\end{multline}
where $\bm{x}\!\in\!\mathbb{R}^{DCT}$, $\bm{S}_{\bm{x}}\!\in\!\mathbb{R}^{DCT\times DC}$, $\bm{x}_1\!\in\!\mathbb{R}^{DC}$, $\bm{S}_{\bm{u}}\!\in\!\mathbb{R}^{DCT\times d(T-1)}$ and $\bm{u}\!\in\!\mathbb{R}^{d(T-1)}$.

\section*{Acknowledgement}
This work has been carried out in the CoLLaboratE project (\url{https://collaborate-project.eu/}), funded by the EU within H2020-DT-FOF-02-2018 under grant agreement 820767 and by the MEMMO project (Memory of Motion, \url{http://www.memmo-project.eu/}), funded by the European Commission's Horizon 2020 Programme (H2020/2018-20) under grant agreement 780684.

\bibliographystyle{spbasic}
\bibliography{./ICRA2019}

\end{document}